# Measurement of Road Traffic Parameters based on Multi-Vehicle Tracking

Kristian Kovačić, Edouard Ivanjko and Niko Jelušić
Department of Intelligent Transportation Systems
Faculty of Transport and Traffic Sciences
University of Zagreb
Email: kristian.kovacic@fpz.hr, edouard.ivanjko@fpz.hr, niko.jelusic@fpz.hr

*Abstract*—Development of computing power and cheap video cameras enabled today's traffic management systems to include more cameras and computer vision based applications for monitoring and control of road transportation systems. Combined with image processing algorithms cameras are used as sensors to measure road traffic parameters like flow volume, origin-destination matrices, classify vehicles, etc. In this paper we propose a system for measurement of road traffic parameters (basic motion model parameters and macro-scopic traffic parameters). The system is based on Local Binary Pattern image features classification with a cascade of Gentle Adaboost classifiers to determine vehicle existence and its location in an image. Additionally, vehicle tracking and counting in a road traffic video is performed by using Extended Kalman Filter and virtual markers. The newly proposed system is compared with a system based on background subtraction. Comparison is performed by means of evaluating execution time and accuracy.

## I. INTRODUCTION

Recent decades can be characterized by a significant increase of the number of road vehicles accompanied by a build-up of road infrastructure. As a consequence, the number of traffic accidents has also increased. Statistics show that there is about $1.2$ million fatalities and $50$ million injuries related to road traffic accidents per year worldwide [1]. Statistics published by the Croatian Bureau of Statistics [2] reveal that in 2011 there were 418 killed and $18,065$ injured persons who were participating in traffic in the Republic of Croatia. In the attempt to reduce the amount of annual traffic accidents, a large number of different systems have been researched and developed. Such systems are part of the road infrastructure (horizontal and vertical signalization, variable message signs, etc.) or road vehicles (various advanced driver support systems). Vehicle manufacturers have implemented systems such as lane detection and lane departure, parking assistants, collision avoidance, adaptive cruise control, etc.

Today's road traffic measurement and management systems most often use computer vision algorithms and complementary metal oxide semiconductor (CMOS) video cameras. This is the consequence of the increase of computing power and decrease of CMOS video camera prices. Cameras are used as a part of the vehicle or road infrastructure. From the obtained video footage, high level road traffic information can be extracted. Such high level information includes incident detection, vehicle classification, origin-destination (OD) matrix estimation, etc. This information needs to be extracted in real-time with high accuracy. Mentioned information is crucial in advanced traffic management systems from the domain of intelligent transportation systems (ITS). Advanced traffic management systems have the goal to stabilize traffic flow and consequently enhance level of service for processes in a road traffic network from various aspects such as road safety, productivity loss, environment pollution, etc.

Basic parameters, which can define a basic motion model, of an individual vehicle are: time $t$ needed to travel a specific distance $s$, current direction $\phi$, velocity $v$, acceleration $a$ and impulse of movement $I$. Additional parameters such as vehicle traffic flow volume $q\left[\frac{veh}{h}\right]$, traffic flow density $g\left[\frac{veh}{km}\right]$, traffic flow velocity $v\left[\frac{km}{h}\right]$, travel time in the traffic flow $t\,[h]$ and headway $S\,[m]$ can be used to define the state of a traffic network [3]. All mentioned parameters can be obtained using only video cameras as sensors.

Road traffic monitoring from the microscopic aspect is based on monitoring basic traffic parameters of each individual vehicle. These parameters are essential for traffic management on a smaller traffic network where mutual influence of vehicles needs to be analyzed. This is crucial to make the proper decision related to traffic management of a traffic network. In larger traffic networks macroscopic traffic parameters are also required for traffic management. With such parameters, physics and influence of each individual vehicle is not analyzed. Instead, traffic is represented by a statistical model where movement of vehicles is described using traffic flow. For a specific traffic flow, certain laws are applied by which it changes through time and space [3].

In this paper, we propose a system for road vehicle detection, tracking and counting. In section II, state-of-the-art (SOTA) approaches for measuring road traffic parameters are explained. Additionally, a similar system proposed in previous research is described. Section III describes the proposed system for vehicle detection. In this section, used image features and classifiers are explained in more detail. In section IV, experimental results are analyzed. We conclude and describe future work of this research in section V.

## II. MEASUREMENT OF ROAD TRAFFIC PARAMETERS USING VIDEO CAMERAS

Measurement of traffic parameters (basic and complex) using video cameras is based on three main steps. First step consists of detecting a vehicle in an image. Result of this step is the location and size of a vehicle in the image. Second stage consists of tracking individual vehicles through multiple consecutive images. In this step, vehicle trajectory can be obtained and analyzed for further computation of traffic parameters. Last step of traffic parameter measurement is vehicle counting by which parameters such as traffic flow volume can be computed.

*A. Problems and basic approaches*

As mentioned previously, there are three main steps in a system for traffic parameters measurement using video cameras. First step, which includes vehicle detection, can be performed using various computer vision methods. One of the commonly used methods is background subtraction. It is based on separating foreground and background segments of an image. This can be performed by comparing two or more consecutive images. When only two images are compared, simple subtraction between two pixels intensities at the same location in the images can be performed. By comparing the subtracted value and a specific threshold value, pixel on the specific location in an image can be classified as a foreground or background pixel [4]. Another approach of a background subtraction method is based on using a Gauss Mixture Model (GMM). GMM is based on computing a probability that certain pixel intensity value at a specific location in an image represents a foreground or background object. If a pixel in the image has probability to be part of the background larger than a specific threshold it is classified as background or otherwise as foreground [5]. Main deficiency of mentioned methods is their incapability to detect static vehicles in an image (e.g. parked vehicles or vehicles in a queue). Methods require a static camera from which the road traffic video is acquired (camera does not change its location and position under influence of wind, vehicle vibrations, etc.).

A similar method where pixels are compared in consecutive images is based on optical flow. Basic work principle of this method is to compute the optical flow of pixels or regions in images. This is achieved by computing an offsets of pixels or regions related to two or more consecutive images. Pixel displacement, direction and velocity are most commonly used as parameters of the optical flow.

System for vehicle detection based on machine learning can use a combination of artificial neural network (ANN), fuzzy logic and genetic algorithm based methods. Today's SOTA approach based on ANN is the deep learning method. Main benefits of deep learning are learning of a multi-step algorithm and representation learning. Another machine learning approach is based on classifiers and hard-coded image features. Both mentioned machine learning approaches require a learning dataset that consists of a large number of positive and negative samples. Positive samples represent images that contain an object and negative samples represent images that do not contain an object which wants to be detected in an image. In the approach based on classifiers and image features, image features are computed for every sample in the learning dataset. Many types of image features exist such as Histogram of Oriented Gradients (HOG), Scale-Invariant Feature Transform (SIFT), Speeded-Up Robust Features (SURF), HAAR, Local Binary Patterns (LBP), etc [6]. After mentioned features are computed for all samples in the learning dataset, learning process of the chosen classifier is performed. After the classifier is learned to perform classification of image features, object detection can be performed. Image features are computed from the new image and separated into two classes: (i) object class; (ii) and non-object class.

*B. State of the art*

To compute any of the previously mentioned road traffic parameters, system needs to perform image processing on videos obtained from road traffic video cameras. As mentioned in the previous section, one of tasks, which needs to be solved, is vehicle detection in an image. Cheng, Lin, Chen, et al. [7] proposed a system which uses a feature extraction and classification method for vehicle detection in an image. System is based on a Support Vector Machine (SVM) classifier, which can perform classification of features into one of two classes defined by Eq. 1. In Eq. 1, $f(x)$ is the result of classification performed by SVM and $K(sv, x)$ is the kernel function which uses a support vector $sv$ and a feature vector $x$ to perform weak classification multiplied by a coefficient $\alpha_n$. The kernel function $K(sv, x)$ is a radial based function. The feature vector $x$ uses Histogram of Oriented Gradient (HOG) image features. Hit rate represents ratio between the number of detected vehicles and ground truth. Hit rate is above $94\%$ when a small number of false positive detection exists [7].

$$\begin{aligned} f(x) &= \alpha_1 K(sv_1, x) + \alpha_2 K(sv_2, x) + \\ &\quad ... + \alpha_m K(sv_m, x) \end{aligned} \quad (1)$$

In the system proposed in [8], vehicle detection and tracking is performed by two separated algorithms. Vehicle detection is performed by a background subtraction method and combination of morphological operations. Benefit of using these methods is algorithm simplicity and its fast execution. Object tracking is based on a Lucas-Kanade pyramid based optical flow algorithm. Overall accuracy of the system in experimental results is $77.9\%$, where object detection accuracy is $81.75\%$ and object tracking accuracy is $95.3\%$ [8].

Problem with both previously mentioned systems is that their accuracy highly depends on camera perspective and scenario situation. If the same system for vehicle detection is used on multiple camera mounting locations (above or beside the road), system parameters need to be adjusted for every location manually. In [9], the system proposed for vehicle detection and tracking tackles the mentioned multi-perspective problem. The system is based on Haar-like features and a cascade of Adaboost classifiers. Learning dataset consists of approximately million samples which are separated into smaller subsets depending on vehicle motion direction. Experimental results show that the hit rate is above $75\%$ when the number of false positives is above $100$ of total $700$.

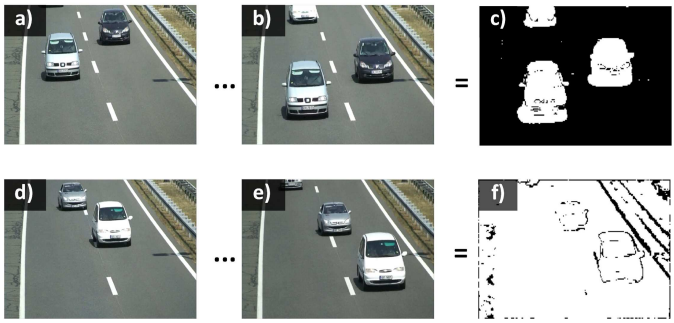

Figure 1. Original sets of images a-b and d-e processed by background subtraction method c and f.

Another problem with mentioned systems is their application in urban scenarios which represents a challenging environment. System proposed in [10] is based on a hybrid image template (HIT). The HIT consists of multiple image patches with various types of features including a sketch, texture, flatness and color. Image sketch patch is based on Gabor wavelet elements, the texture and flatness patch are based on Gabor filter [11] and color is extracted through histogram in HSV color space. A probability model of the HIT is defined for assisting vehicle detection. Vehicle detection is performed in three stages through the SUM-MAX procedure where in each stage a local image region with the maximal score is selected as a vehicle candidate. If this score is greater than the learned threshold, candidate is viewed as a detected vehicle object. The experimental results show that the HIT preferably extracts meaningful features including sketch, texture, flatness, and color for vehicle object, and improves the vehicle detection accuracy in complex urban traffic conditions. The proposed system adapts to slight vehicle deformation also.

*C. Previous research*

In previous research, authors proposed a system for vehicle detection based on the background subtraction method which is described in [18]. As a result of the background subtraction method, each pixel in the image is classified as 1 if it is a part of the foreground (moving vehicles) or 0 if it is a part of the background (static object). Main disadvantage of the background subtraction method is shown in Fig. 1. From the set of images between 1a-1b and 1d-1e, a background model is created and subtracted from the newest image, which result is shown in c and f. In the given examples, the result of background subtraction given in the image c has a small number of incorrectly classified pixels. In the image f most of pixels are incorrectly classified due to sudden light change in the whole image. This event most often occurs as a result of a large bright object (e.g. truck) passing the scene or when the camera changes its position in an event of strong wind.

### III. APPROACH BASED ON CLASSIFIER LEARNING

System proposed in this paper consists of three main components. First component performs vehicle detection, second component tracks detected vehicles in the image plane and third component performs vehicle counting in order to obtain traffic parameters.

*A. Vehicle detection*

The proposed system in this paper is based on a method that extracts Multi-Scale LBP (MB-LBP) features from an image as described in [12] and performs classification with the Gentle Adaboost (GAB) algorithm explained in [13] and [14]. Methods are implemented in OpenCV library with CUDA support. Before vehicle detection can be performed in an image, a classifier needs to be learned to classify vehicle MB-LBP features in the image. The learning process is performed with a learning dataset consisting of two subsets of images:

- Images which contain vehicles (positive samples);
- Images which do not contain vehicles (negative samples).

From the individual subsets, MB-LBP features are extracted. The classifier is learned to separate specific features into two classes (with and without vehicles). Three versions of dataset are used to learn the classifier as described in continuation of this paper.

*1) Image features:* MB-LBP features represent modified version of LBP features that are originally described in [15]. In [15], LBP features are used for face detection, however they can be applied on different type of objects such as vehicles [14]. MB-LBP features are similar to LBP features and they are both described in continuation of this subsection. First step of LBP feature extraction is comparison of the specific pixel in an image with each of its 8 adjacent pixels. If an adjacent pixel intensity is greater than or equal to the middle pixel, cell with bit 1 is used in a binary string, or bit 0 otherwise as shown in Fig. 2a. Binary string consists of 8 bits where each bit is set to previously mentioned result of the adjacent pixel comparison. In Fig 2a, bits are stored clockwise into the binary string, where the binary string value would be 11010011. This procedure is repeated for every pixel in the region for which LBP feature extraction needs to be performed. MB-LBP feature extraction is based on multi-scale cells where each cell consists of more than one pixel as shown in Fig. 2b. By this method, MB-LBP features can describe a texture in the image in micro-scopic and macro-scopic level. After LBP or MB-LBP features are extracted for each pixel, a histogram is computed based on binary string values. In a LBP feature histogram, every combination of a uniform LBP feature is stored in a separate histogram bin with one additional bin space for all features that are not uniform. Uniform LBP features denote all binary strings which contain no more than two transitions between bit 0 and 1. In a MB-LBP feature histogram instead of using uniform features, first 63 features with the highest rank (occurrence) are stored in histogram bins with one additional histogram bin which summarizes all features which are below first 63 features with the highest rank [16].

*2) Classifier:* In the proposed system, cascade of classifiers is used for classification process. Decision stump is used as a weak classifier and Gentle Adaboost is used as a strong classifier. Decision stump as a weak classifier computes the hypothesis $h_t(x)$ with accuracy a slightly higher than random guessing. Multiple weak classifiers combined together in the strong classifier allow computation of the strong hypothesis with high accuracy. A weak classifier can be defined with Eq. 2, where $h_t(x)$ is a weak hypothesis, $x$ is a feature vector and $threshold$ is a constant value used for comparison with a feature $x_i$ in the feature vector $x$. The $threshold$ is determined in classifier learning process. Strong classifier is defined by Eq. 3, where $H(x)$ is a strong hypothesis, $\alpha_t$ is a weight factor computed in a learning step of classifier [13].

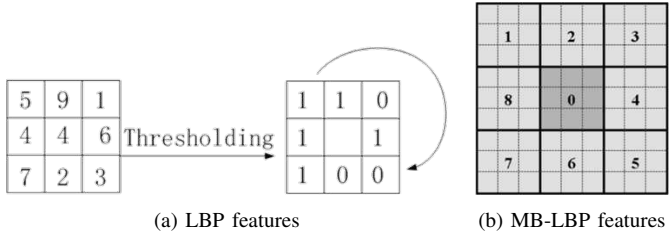

(a) LBP features  (b) MB-LBP features

Figure 2. Principle of feature extraction [16].

Table 1: Number of samples in datasets.

| Dataset created by | Number of positive samples | Number of negative samples | Total number of samples |
|---|---|---|---|
| Faculty of Transport and Traffic Sciences (FTTS) | 4,298 | 1,930 | 6,228 |
| Toyota Motor Europe (TME) Motorway dataset [17] | 25,155 | 26,321 | 51,476 |
| Combination of the both datasets | 29,453 | 28,251 | 57,704 |

$$h_t(x) = \begin{cases} -1, & if\ x_i < threshold \\ 1, & if\ x_i \geq threshold \end{cases} \quad (2)$$

$$H(x) = sign\left(\sum_{t=1}^{T} \alpha_t h_t(x)\right) \quad (3)$$

Cascade classifier consists of multiple stages, where in each stage is one of the strong classifiers. In the first stage, strong classifier computes $H(x)$ based on a small number of features in the feature vector $x$. In each following stage, a strong classifier uses an increased number of features in the feature vector $x$, obtaining more and more accurate hypothesis. This process reduces execution time of classification algorithm.

*3) Classifier learning process:* Before vehicle detection can be performed in the image, a classifier needs to be learned to classify vehicle LBP features in the image. The learning process is performed using a learning dataset with the properties described above.

- Images which contain vehicles (positive samples);
- Images which do not contain vehicles (negative samples).

From the individual subsets, LBP features are extracted. The classifier is learned to separate specific features into two classes (with and without vehicles). In this paper, three versions of datasets described in Tab. 1 are used to learn the classifier. The dataset created by the Faculty of Transport and Traffic Sciences is made from video footage of a road near the city of Zagreb obtained by a full HD camera with resolution $1920x1080$. Dataset contains images of a vehicle from front-side and back-side perspective obtained using a camera mounted above the road. Dataset made by TME [17] contains images of a vehicle only in back-side perspective obtained from a camera mounted in a vehicle.

*B. Vehicle tracking*

Trajectory computation of a moving object in a set of consecutive images is performed by EKF. Each detected object in a scene is separately tracked by EKF. An EKF state model is based on a vector $x$ defined in Eq. 4, where $x_x$ and $x_y$ represent location of the moving object on the x and y axis, $x_v$ is the velocity and $x_a$ is the acceleration of the moving object over an image plane, $x_\phi$ and $x_\omega$ are the object direction and change of direction in a time interval $t$. Prediction of a future model state is performed with Eq. 5, where $x_{k|k-1}$ is a predicted state model vector in the step $k$ based on a state model vector $x_{k-1|k-1}$ from the previous step $k-1$ and $t$ is the time interval between the steps $k-1$ and $k$.

$$x = \begin{bmatrix} x_x \\ x_y \\ x_v \\ x_a \\ x_\phi \\ x_\omega \end{bmatrix} \quad (4)$$

$$x_{k|k-1} = f(x_{k-1|k-1}, t) \quad (5)$$

After the prediction step has been finished, an update step is performed which is described in more details in [18]. For measurement step in EKF, data from the applied vehicle detection method (location, size) is used.

*C. Vehicle counting*

The last step in the proposed system is vehicle counting. Vehicle counting is based on virtual markers $M_1$ and $M_2$ shown in Fig. 3 as rectangles 0 and 1. Virtual markers are located in the bottom part of the image in order to detect that an object is leaving the scene. Detection method is based on checking overlaps between each moving object and virtual marker. When the background subtraction method is used, a vehicle counter will be increased by 1 if following requirements are fulfilled:

- The total number of images in which an object was present in the scene is greater than *TFC*;
- When the EKF is used for vehicle tracking, vehicle is overlapping with a virtual marker if the direction of vehicle movement is inside a specific interval $[\phi_{min}, \phi_{max}]$, where $\phi_{min}$ and $\phi_{max}$ are set by Eqs. 6-7.

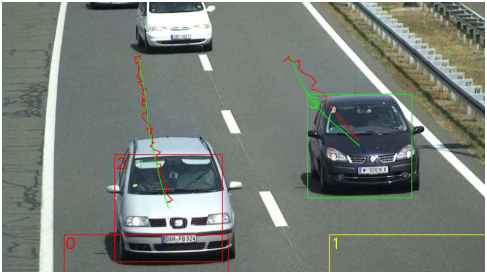

Figure 3. Testing of vehicle detection, tracking and counting algorithms.

Table 2: Experimental results for the background subtraction method.

| Image resolution | Method parameters | | Counted vehicles | |
|---|---|---|---|---|
| | th | TFC | FP / FN / GT | A [%] |
| Without EKF | | | | |
| $240 \times 135$ | 10 | 15 | 4 / 2 / 133 | 95 |
| $240 \times 135$ | 10 | 30 | 2 / 6 / 133 | 94 |
| $240 \times 135$ | 8 | 15 | 5 / 6 / 133 | 92 |
| $240 \times 135$ | 8 | 20 | 5 / 8 / 133 | 90 |
| With EKF | | | | |
| $240 \times 135$ | 10 | 10 | 4 / 4 / 133 | 94 |
| $240 \times 135$ | 10 | 30 | 3 / 6 / 133 | 93 |
| $240 \times 135$ | 8 | 10 | 4 / 6 / 133 | 92 |
| $240 \times 135$ | 8 | 30 | 4 / 8 / 133 | 91 |

$$\phi_{min} = \left(\frac{1}{8} + 1\right)\pi [rad] \quad (6)$$

$$\phi_{max} = \left(\frac{7}{8} + 1\right)\pi [rad] \quad (7)$$

When the feature classification method is used, a vehicle counter will be increased by 1 if following requirements are fulfilled:

- The total number of images in which an object was present in the scene is greater than $TFC$;
- The total travelled distance of an object in the image plane is larger than $\frac{1}{5}max(w,h)$, where $w$ is the image width $[px]$ and $h$ is the image height $[px]$.

IV. EXPERIMENTAL RESULTS

In this section experimental results are given which show comparison between previously mentioned methods for vehicle detection and tracking from aspects of accuracy and execution time. Used video footage is recorded by a camera mounted above the bypass of the city of Zagreb at the location Ivanja Reka. The camera records the vehicle front side and therefore combination of FTTS and TME dataset is used. Results of testing the background subtraction method are given in Tab. 2, where image resolution of $240 \times 135\ [px]$ is used in vehicle detection, *FP* is the number of false positive detections, *FN* is the number of false negative detections and *GT* is the groundtruth data, *TFC* is the minimum number of images in which an object needs to be detected to be classified as vehicle. Testing of the background subtraction method is performed with and without EKF as noted in Tab. 2 on the whole video footage. Accuracy *A* is defined by Eq. 8. Average execution time of background subtraction method in all testings has the same value of 16 *[ms]*. Experimental results of the feature classification method are given in Tab. 3, where *S* is the number of stages in a cascade classifier, *MHR* (Minimum Hit Rate) is the minimum ratio between the number of detected objects and the total number of objects which have to pass into a next learning stage of a cascade classifier and *MCC* (Minimum Cluster Count) is the minimum number of individual clusters detected by the feature classifier which an object (vehicle) needs to contain. Testing of the proposed method is performed with and without EKF same as in the case of the background subtraction method. Execution time in *[ms]* is given in Fig. 4 where the fastest measurement regarding the execution time was approximately 53 *[ms]*. In tested methods, a false negative detections occur if a virtual marker has not been activated due to a vehicle passing through the scene. This occurs if a vehicle is not detected on every image and the total number of images, on which vehicle was present, is less than *TFC* parameter. A false positive detection occurs if a vehicle is wrongly classified into two or more vehicles in the scene.

$$\left(1 - \frac{FP + FN}{GT}\right) \times 100\% \quad (8)$$

The system based on the background subtraction method was tested on all combinations of the following method parameters:

- Image resolution used in the detection process = $\{240 \times 135, 480 \times 270, 960 \times 540\}$;
- Threshold value $th = \left\{\begin{array}{l}5, 8, 10, 15, 20, 25,\\ 30, 35, 40, 45, 50\end{array}\right\}$;
- $TFC = \{10, 15, 20, 30\}$;
- $EKF = \{used, notused\}$.

Testing values for parameters of the system based on the feature classification method where:

- Image resolution used in the detection process = $\{240 \times 135, 480 \times 270, 960 \times 540\}$;
- $TFC = \{3, 8, 15\}$;
- $MHR = \{0.99, 0.9925, 0.995, 0.9975, 0.999875\}$;
- The number of stages in a cascade $N = \{5, 10, 20\}$;
- $EKF = \{used, notused\}$.

V. CONCLUSION & FUTURE WORK

In this paper, experimental results of two systems for vehicle detection and tracking in road traffic video footage are compared. Testings are performed with various system parameter values where maximum achieved vehicle counting accuracy is $95\%$. From obtained experimental results, it can be concluded that with the feature classification method similar accuracy can be obtained as with the background subtraction method. System based on the background subtraction method has the advantage that it can detect any moving object in the video footage. As a consequence, determination of system parameters is a simple process. Main disadvantage of this method is its sensitivity to camera vibrations and sudden lighting changes in an image which can cause increase of false positive and false negative detections.

Table 3: Experimental results for the feature classification method.

| Image resolution | Method parameters | | | | Counted vehicles | A [%] |
|---|---|---|---|---|---|---|
| | S | MHR | MCC | TFC | FP / FN / GT | |
| Without EKF | | | | | | |
| 240 × 135 | 20 | 0.995 | 2 | 8 | 0 / 6 / 133 | 95 |
| 240 × 135 | 20 | 0.999875 | 5 | 3 | 0 / 7 / 133 | 95 |
| 240 × 135 | 20 | 0.999875 | 5 | 15 | 0 / 7 / 133 | 95 |
| 240 × 135 | 20 | 0.995 | 2 | 3 | 3 / 5 / 133 | 94 |
| 480 × 270 | 20 | 0.995 | 5 | 3 | 3 / 6 / 133 | 93 |
| With EKF | | | | | | |
| 240 × 135 | 20 | 0.995 | 2 | 3 | 0 / 9 / 133 | 93 |
| 240 × 135 | 20 | 0.995 | 2 | 8 | 0 / 9 / 133 | 93 |
| 240 × 135 | 20 | 0.995 | 2 | 15 | 0 / 9 / 133 | 93 |
| 240 × 135 | 20 | 0.999875 | 5 | 3 | 0 / 9 / 133 | 93 |
| 240 × 135 | 20 | 0.999875 | 5 | 8 | 0 / 9 / 133 | 93 |

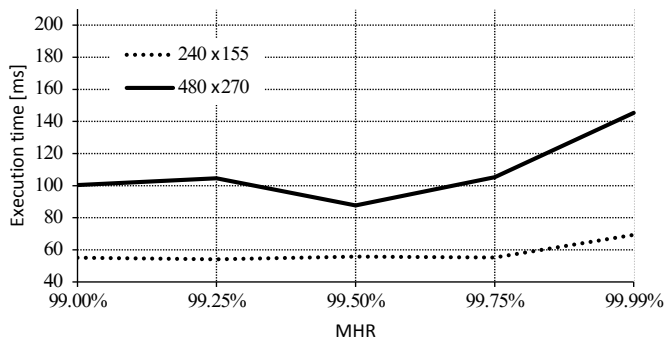

Figure 4. Execution time of the proposed system based on the feature classification method.

When a feature classification method is used, the system is robust to camera vibrations and sudden lighting changes in the image, and its accuracy keeps the same value. Disadvantage of this method is the requirement for a classifier learning process and a large dataset in order to perform accurate object detection. This requirement consumes a lot of time spent on preparation of a dataset and the classifier learning process. Mentioned method increases the number of system parameters, where determination of initial values of parameters is a more complex process which requires a lot of testings and system analysis. Another disadvantage of the newly proposed feature classification based method is its execution time which is substantially longer than in the case of the background subtraction method.

Future work of this research would consists of expanding the current dataset with side images of vehicles, images of heavy vehicles (trucks) and motorcycles. This would allow detection of vehicles in various poses. Optimization of currently developed method from an aspect of execution time would need to be examined in order to provide real-time vehicle detection. Additional analysis of the system accuracy when other image features are used such as SURF, HOG or SIFT will be made also.

ACKNOWLEDGMENT

This work has been supported by the IPA2007/HR/16IPO/001-040514 project "VISTA - Computer Vision Innovations for Safe Traffic" which is co-financed by the European Union from the European Regional and Development Fund and by the EU COST action TU1102 - "Towards Autonomic Road Transport Support Systems". The authors also thank Mario Muštra for his helpful suggestions and comments.

REFERENCES

[1] C. Hughes, R. O'Malley, D. O'Cualain, M. Glavin and E. Jones: "New Trends and Developments in Automotive System Engineering: Chapter Trends Towards Automotive Electronic Vision Systems for Mitigation of Accidents in Safety Critical Situations" - InTech, 2011., pp. 493-512

[2] E. Omerzo, S. Kos, A. Veledar, K. Šakić Pokrivač, L. Pilat, M. Pavišić, A. Belošević, and Š. V. Njegovan: "Deaths in Traffic Accidents" - Croatian Bureau of Statistics, 2012.

[3] G. Kos and I. Dadić: "Theory and organization of traffic flows" - Faculty of Transport and Traffic Sciences, University of Zagreb, 2007.

[4] M. Piccardi: "Background subtraction techniques: a review" - Proceedings of IEEE Conference SMC, Vol. 4, The Hague, October, 2004., pp. 3099-3104

[5] D. Arita, T. Tanaka, A. Shimada and R. I. Taniguchi: "A fast algorithm for adaptive background model construction using parzen density estimation" - Proceedings of IEEE Conference AVSS, London, September, 2007., pp. 528–533

[6] X. Jiang: "Feature extraction for image recognition and computer vision" - Proceedings of IEEE Conference ICCSIT, Beijing, August, 2009., pp. 1-15

[7] K.-M. Cheng, C.-Y. Lin, Y.-C. Chen, T.-F. Su, S.-H. Lai and J.-K. Lee: "Design of Vehicle Detection Methods With OpenCL Programming on Multi-Core Systems" - Proceedings of IEEE Conference ESTIMedia, Montreal, October, 2013., pp. 88-95

[8] K. Kiratiratanapruk and S. Siddhichai: "Vehicle Detection and Tracking for Traffic Monitoring System" - Proceedings of IEEE Conferenece TENCON, Hong Kong, November, 2006., pp. 1-4

[9] R. Feris, S. Pankanti and B. Siddique: "Learning Detectors from Large Datasets for Object Retrieval in Video Surveillance" - Proceedings of IEEE Conference ICME, Melbourne, July, 2012., pp. 284-289

[10] Y. Li, B. Li, B. Tian, F. Zhu, G. Xiong and K. Wang: "Vehicle Detection based on the Deformable Hybrid Image Template" - Proceedings of IEEE Conference ICVES, Dongguan, July, 2013., pp. 114-118

[11] V. S. N. Prasad and J. Domke: "Gabor Filter Visualization" - Technical report, University of Maryland, 2005.

[12] M. Kurt, B. Kir and O. Urhan: "Local binary pattern based fast digital image stabilization" - IEEE Signal Processing Letters, IEEE Journals & Magazines, Vol. 22, No. 3, 2015., pp. 341-345

[13] G. Lemaître and M. Radojević: "Directed reading: Boosting algorithms" - Heriot-Watt University, Universitat de Girona, Universite de Bourgogne, December, 2009.

[14] H. Liang, G. Teodoro, H. Ling, E. Blasch, G. Chen and L. Bai: "Multiple kernel learning for vehicle detection in wide area motion imagery" - Proceedings of IEEE Conference FUSION, Singapore, July, 2012., pp. 1629–1636

[15] T. Ojala, M. Pietikäinen and D. Harwood: "A comparative study of texture measures with classification based on feature distributions" - Pattern Recognition, Vol. 29, No. 1, January, 1996, pp. 51-59

[16] S. Liao, X. Zhu, Z. Lei, L. Zhang and Stan. Z. Li: "Learning Multi-scale Block Local Binary Patterns for Face Recognition" - Proceedings of IEEE Conference ICB, August, 2007., pp. 828-837

[17] C. Caraffi, T. Vojir, J. Trefny, J. Sochman and J. Matas: "A system for real-time detection and tracking of vehicles from a single car-mounted camera" - Proceedings of IEEE Conference on Intelligent Transportation Systems, Anchorage, September, 2012., pp. 975-982

[18] K. Kovačić, E. Ivanjko and H. Gold: "Real Time Vehicle Trajectory Estimation on Multiple Lanes" - Proceedings of 3rd CCVW2014, Zagreb, September, 2014., pp. 21-26